\documentclass[runningheads]{llncs}
\usepackage{tabularx,ragged2e,booktabs,makecell,authblk, float}
\usepackage{array}
\renewcommand{\arraystretch}{0.9}
\usepackage[backend=biber,style=numeric,sorting=none,maxcitenames=2]{biblatex}
\usepackage[T1]{fontenc}
\usepackage{graphicx}
\begin{document}
\title{Context-Aware Classification and Grading of Sensitive Information in Online Conversational Health Data}
\titlerunning{Context-Aware Classification and Grading}
%
\author{Yiwei Yan \and Guanfeng Liu}
\institute{School of Computing, Macquarie University, Australia
}

\authorrunning{Y. Yan et al.}
%

%
\maketitle             
\begin{abstract}
Online medical consultations contain sensitive health information whose privacy implications depend not only on the entities mentioned but also on how those entities are described in context. Existing classification and grading approaches often map health-information entities directly to predefined sensitivity levels, potentially overlooking whether a condition is confirmed, suspected, negated, hypothetical, or merely planned for investigation. In this study, we formulate sensitive-information grading in online medical dialogues as a context-aware evaluation task. We develop a standard-informed operational framework that incorporates assertion status, experiencer, test-result status, and information granularity. We further design a naturalistic evaluation setting together with contrastive cases that minimally alter negation, uncertainty, experiencer, or granularity, and compare large language models under mention-only and full-context conditions. The study aims to quantify the contribution of contextual information to sensitivity grading and to characterize safety-critical over- and under-grading errors. Our framework provides a reproducible basis for evaluating whether LLMs can distinguish sensitive entity mentions from contextually established sensitive disclosures.

\end{abstract}
\section{Introduction}
The pervasive digitization of health data has propelled healthcare delivery and technological innovation \cite{bajwa2021artificial}. Meanwhile, significant privacy concerns have been raised due to the high sensitivity and complexity of health data, which requires a thoughtful balance between innovation and patient privacy security \parencites{wan2016characteristics, faridoon2024healthcare}. This concern is shared globally; the Health Insurance Portability and Accountability Act (HIPAA) mandates that 18 types of re-identifiable information must be removed from medical records before sharing. The General Data Protection Regulation (GDPR) in the European Union requires the anonymization or pseudonymization of re-identifiable personal data \parencites{sweeney2017re, protection2018general}. Specifically, China's Data Security Law, Cybersecurity Law, and Personal Information Protection Law establish the data graded protection obligations and define key concepts such as ``important data'' and ``sensitive personal information'' \parencites{chen2021understanding,creemers2023cybersecurity,calzada2022citizens}. For cross-industry implementation, GB/T 43697-2024 provides general rules, principles, and processes for data classification and grading, which are aligned with data attributes and sensitivity \cite{china2024data}.

In healthcare, the national standard ``Information Security Technology – Guide for Health Data Security'' (GB/T 39725-2020) sets scenario-oriented rules that divide health data into six categories (personal attributes, health status, medical applications, medical payment, health resources and public health) and five ascending sensitivity levels (1-5), lacking detailed and operational criteria for classification and grading \cite{standardization2020information}. Moreover, GB/T 39725-2020 focuses primarily on electronic medical record (EMR) security and offers limited guidance for online conversational health data. With the rise of Internet hospitals in China, healthcare delivery has been shifted from in-person visits to online consultations. The resulting online health data preserve original and conversational interactions between patients and doctors, and commonly contain large amounts of protected health information (PHI). Accordingly, data classification and grading tailored to online medical services are urgently needed, with principled rules and practical procedures.

Natural language processing (NLP) is widely used for information-extraction tasks to discover critical knowledge in unstructured data \cite{malmasi2018extracting}. In practice, workflows rely on manual annotation to build gold-standard datasets and on named-entity recognition (NER) systems for automated entity identification. However, rule- and metadata-based NER depends on fixed rules and labeled data, limiting its capacity to capture complex contextual relationships and generalize to new domains \cite{ghaffari2024framework}. By contrast, accurate classification and grading critically depends on context-aware understanding. For instance, data related to special diseases (e.g., sexually transmitted infections (STD)) is graded as level 5 under GB/T 39725-2020 \cite{standardization2020information}. In fact, patients may simply inquire about special diseases in online consultations, without actually having them. Assigning ``level 5'' labels to suspected or explicitly ruled-out mentions of special diseases is inappropriate. Recently, large language models (LLMs) have shown remarkable ability to process diverse text and capture complex entity relationships through deep context understanding and pre-trained knowledge. For instance, OpenAI’s ChatGPT achieved passing grades in the United States Medical Licensing Examinations, evidencing its robust medical knowledge and reasoning \parencites{kung2023performance,nori2023capabilities}. Recent researches have demonstrated the potential of LLMs to identify private information in medical data. DeID-GPT, a medical text de-identification model based on GPT-4, has shown high accuracy and reliability in masking private information within unstructured medical text \cite{liu2023deid}. The LLM-Anonymizer, developed using Llama-3 70B, demonstrates effective performance in removing personal identifiable information from text \cite{wiest2024anonymizing}. 

Motivated by the strong capabilities of LLMs, this research focuses on the context-aware study of LLM-based methods for data classification and grading, especially for online conversational health data. Our main contributions are as follows.
\begin{enumerate}
    \item We formulate sensitive-information grading as a context-dependent task rather than a direct mapping from entity mentions to sensitivity levels.
    \item We develop a standard-informed, assertion-aware operational grading framework that explicitly distinguishes confirmed, suspected, negated, planned, and hypothetical medical information.
    \item We propose a naturalistic and contrastive evaluation framework to examine whether current LLMs respond appropriately to changes in assertion, experiencer, and information granularity.
\end{enumerate}

\section{Related Work and Regulatory Background}

\subsection{Health Data Classification and Grading}

GB/T 39725-2020, together with China's broader data-security and personal-information protection framework, provides the regulatory basis for health-data classification and grading. The Standard defines major categories of health data and organizes them according to increasing sensitivity and potential impact. The relevant classification and grading principles are summarized below.

\subsubsection{Health data classification}

The ``Standard'' clearly defines ``health data'' as ``personal health data and health-related electronic data obtained after processing personal health data'', which can be divided into six categories: personal attributes, health status, medical applications, medical payment, health resources and public health (Table ~\ref{tab:health_data_categories}).

\begin{table*}[htbp]
\centering
\caption{Definition and scope of health data categories}
\label{tab:health_data_categories}
\small
\renewcommand{\arraystretch}{1.2}
\begin{tabularx}{\textwidth}{
    >{\raggedright\arraybackslash}p{0.17\textwidth}
    >{\raggedright\arraybackslash}p{0.27\textwidth}
    >{\raggedright\arraybackslash}X}
\toprule
\textbf{Health data classification} &
\textbf{Definition} &
\textbf{Scope} \\
\midrule

Personal attribute data &
Information that can identify a specific individual, either on its own or when combined with other information. &
Demographic information, personal identifiers, contact information, biometric information, and health-monitoring device identifiers. \\
\midrule

Health status data &
Information that reflects an individual's health condition or is closely related to it. &
Chief complaints, medical history, symptoms and signs, laboratory and examination data, family history, lifestyle information, and genomic or other biological data. \\
\midrule

Medical application data &
Information that reflects the provision and use of medical services, including outpatient care, hospitalization, and discharge. &
Medical records, prescriptions, examination and laboratory reports, medication information, surgical and anesthesia records, nursing records, and referral or discharge records. \\
\midrule

Medical payment data &
Information related to expenses incurred for medical or insurance services. &
Medical transactions, payment information, insurance status, and insurance-related records. \\
\midrule

Health resource data &
Information that reflects the capabilities and characteristics of health service personnel, health plans, and health systems. &
Hospital basic information, healthcare resources, personnel information, and hospital operation data. \\
\midrule

Public health data &
Information related to public health matters affecting populations at the national or regional level. &
Environmental health, disease surveillance, epidemic information, disease prevention, and population-level birth and mortality data. \\
\bottomrule

\end{tabularx}
\end{table*}

\subsubsection{Health data grading}
Based on the degree of importance and risk of data, as well as the potential damage or impact on individuals if leaked, it is recommended that health data be classified into the following five levels, each corresponding to different restrictions and control measures:
\begin{enumerate}
    \item \textbf{Level 1 (Public data)}: Data that can be directly available to the public on the Internet. This type of data can be fully utilized in public access. 
    \item \textbf{Level 2 (General sensitivity data)}: Data that cannot be identified. This data can be accessed and used within a relatively wide scope, such as for research and analysis by doctors after application and approval.
    \item \textbf{Level 3 (Relatively high sensitivity data)}: Data that may still be re-identified after partial de-identification. This data can be accessed and used within a medium scope, such as by an authorized project team.
    \item \textbf{Level 4 (High sensitivity data)}: Data that can be directly identified to an individual. This data can be accessed and used within a relatively small scope, such as by relevant medical staff.
    \item \textbf{Level 5 (Special disease data)}: Data detailing specific diseases (e.g., HIV, STDs). This data can be accessed and used only within a very small scope and under strictly limited conditions, such as by attending healthcare staff, and is subject to strict controls.
\end{enumerate}

Additionally, the ‘Standard’ explicitly provides some examples for privacy risk grading. For personal attribute data, Level 3 includes the date of birth, date of consultation, date of examination, date of treatment, date of hospitalization, age, work unit, etc. Level 4 includes the name, ID number (such as ID card/driver's license), phone number, fax, email, health insurance number, medical record file number, account, biometric information (like fingerprints and voice not related to the purpose of the application), photo, hobbies, and beliefs. 

For medical application data, Level 2 includes the name of the test, hospital, and department of consultation. Level 3 includes summary report-type information such as test and examination reports, surgical records, and discharge summaries. Level 4 includes detailed hospitalization history., and Level 5 includes information about special diseases.

GB/T 39725-2020 provides an overarching framework for classifying health data into six categories and five sensitivity levels, together with representative examples for several data types. However, it does not provide an exhaustive entity-level mapping for free-text medical dialogues, nor does it explicitly specify how contextual factors such as negation, uncertainty, experiencer, or result status should affect sensitivity grading. We therefore treat the Standard as the basis for a standard-informed operational grading framework, described in Section~\ref{sec:grading_framework}.

\subsection{LLMs for Clinical Privacy and Sensitive Information}
In the literature, recent studies have increasingly explored the use of LLMs for privacy protection in healthcare data. 

\paragraph{\textbf{Medical Text De-identification.}} LLMs have been used to anonymize electronic health records (EHRs). \emph{DeID-GPT} uses GPT-4 to reliably mask PHIs in unstructured medical text, achieving high precision in de-identification tasks \cite{liu2023deid}. Similarly, the \emph{LLM-Anonymizer} based on LLaMA-3 70B demonstrates the feasibility of local deployment for privacy-preserving anonymization in clinical settings \cite{wiest2024anonymizing}. As a framework, \emph{RedactOR} combines rules with LLM inference to handle both textual and audio clinical data \cite{heo2023redactor}. These efforts highlight the potential of LLMs to outperform traditional rule-based or NER-based systems by capturing nuanced context and reducing annotation burdens.

\paragraph{\textbf{Utility-Preserving Anonymization.}} Beyond de-identification, several studies address the challenge of balancing privacy with data utility. Yang et al. proposed a robust anonymization framework where LLMs act as both privacy and utility assessors, optimizing replacements to preserve downstream task performance \cite{yang2024robust}. Kim et al. introduced \emph{SEAL}, a self-refining anonymization model trained via adversarial distillation, enabling effective anonymization while transferring capabilities to smaller models \cite{kim2025self}. Sarkar et al. explored hybrid approaches that integrate synthetic text generation with selective anonymization, providing more flexible privacy–utility trade-offs for clinical note sharing \cite{sarkar2025not}.

\paragraph{\textbf{Privacy Risks.}} In addition to protective mechanisms, researchers have examined the risks LLMs themselves introduce. Staab et al. argued that privacy violations extend beyond memorization, as LLMs may infer sensitive attributes through contextual reasoning even when such attributes are not explicitly present \cite{staab2023beyond}. This line of work underscores the limitations of current anonymization strategies and calls for more robust safeguards against inference-based privacy leakage. Furthermore, \emph{DIRI} is an LLM used to re-identify the patient in the dataset \cite{morris2025diri}. Xu et al. focus on the security of biomedical multimodal LLMs using machine unlearning \cite{xu2025learning}.

\section{Methods}
\paragraph{Problem Setup.}
We consider a set of $N$ target mention--context instances extracted from
online conversational health data. Each instance contains a consultation
identifier, a target span, a reference information type/subtype, and a
protocol-based reference grade. Formally, an instance is represented as
\[
x=(m,t,c),
\]
where $m$ denotes the target mention or span, $t$ denotes its reference
information type or subtype, and $c$ denotes the surrounding dialogue
context. The model produces
\[
y=(a,e,p,g),
\]
where $a$ is assertion status, $e$ is experiencer, $p$ is applicability,
and $g$ is the sensitivity grade. Entity extraction is intentionally
excluded from the primary task so that NER errors do not confound the
evaluation of contextual grading ability.

\subsection{Standard-Informed Operational Grading Framework}
\label{sec:grading_framework}

Based on these regulatory principles, we developed a standard-informed operational grading framework for the information types examined in this study (Table~\ref{tab:sensitivity_grading_rules}). The framework distinguishes standard-explicit or standard-informed mappings from study-defined contextual rules, particularly for assertion status, experiencer, result status, and information granularity.

\begin{table}[h]
    \centering
    \caption{Operational sensitivity grading rules for online conversational health data in this study}
    \label{tab:sensitivity_grading_rules}
    \begin{tabular}{lll}
        \toprule
        \textbf{Information type} &
        \textbf{Contextual condition / granularity} &
        \textbf{Reference grade} \\
        \midrule
        Special disease & Confirmed, specific individual & L5 \\
        Special disease & Suspected & L2 \\
        Special disease & Negated & L2 \\
        Special disease & Hypothetical / planned & L2 \\
        Special disease & Generic / no individual & N/A \\
        \addlinespace
        Sensitive test result & Confirmatory positive & L5 \\
        Test result & Negative / indeterminate & L3 \\
        Test name & Mentioned / planned & L2 \\
        \addlinespace
        Full name & --- & L4 \\
        Surname & --- & L3 \\
        Exact date & --- & L3 \\
        Month / year & --- & L2 \\
        Detailed address & --- & L3 \\
        Coarse address & --- & L2 \\
        \bottomrule
    \end{tabular}
\end{table}

In the present study, the core special-disease lexicon is restricted to
HIV/AIDS, syphilis, gonorrhea, genital warts, and genital herpes.
Other disease groups are not automatically treated as special diseases
unless explicitly covered by the frozen annotation protocol.

\subsection{Naturalistic Evaluation}

The naturalistic evaluation set is designed to include approximately 500 target mention--context instances drawn from at least 350 distinct online medical consultations. Consultations are selected using a predefined random or systematic sampling procedure with a fixed random seed. Eligible sensitive-information mentions are then identified according to the frozen taxonomy and operational grading framework. Multiple target mentions may be extracted from the same consultation, while all analyses preserve consultation-level grouping to avoid information leakage across samples.

To maintain the natural distribution of sensitive information, samples are not enriched on the basis of model outputs or predefined difficult keywords, such as HIV or syphilis. If the initial sample contains fewer than 25 confirmed Level~4--5 instances, sampling may be extended using the same predefined procedure, up to a maximum of 700 target mention--context instances. Each instance is assigned protocol-based reference annotations for assertion status, experiencer, applicability, and sensitivity grade before model evaluation.

\subsection{Contrastive Evaluation}

The contrastive challenge set is designed to contain 300 minimal pairs. Each pair contains two instances with the same target span and reference information type/subtype, while only one predefined contextual factor is altered. This design enables a controlled assessment of whether LLMs appropriately change or preserve sensitivity grades in response to subtle contextual differences.

The challenge set covers four contextual phenomena: 100 negation pairs, 100 uncertainty pairs, 50 experiencer pairs, and 50 granularity pairs. Negation and uncertainty are emphasized because assertion status is central to distinguishing confirmed sensitive disclosures from non-confirmed mentions. Experiencer pairs examine whether models correctly identify whose information is being described, while granularity pairs assess whether models respond appropriately to differences in information specificity.

Representative examples include:
\begin{enumerate}
    \item \textbf{Negation}: 
    A: The patient has been diagnosed with syphilis. \\
    B: Syphilis has been ruled out in the patient.

    \item \textbf{Uncertainty}: 
    A: The patient has confirmed HIV infection. \\
    B: The patient is suspected of having HIV infection.

    \item \textbf{Experiencer}: 
    A: I have tested positive. \\
    B: My mother has tested positive.

    \item \textbf{Granularity}: 
    A: Wang Xiaoming. \\
    B: Dr.~Wang / a patient surnamed Wang.
\end{enumerate}

The two members of each pair are evaluated independently and presented to the model in randomized order so that the model is not informed of their paired relationship. Each pair is reviewed to ensure that no unintended differences in disease severity, experiencer, temporal status, or result status are introduced beyond the intended contrastive factor.

\section{Experiments}
\subsection{Experiment Settings}
\subsubsection{1) Dataset}
\paragraph{MedDialog-CN-2020} We benchmark our method on MedDialog-CN (Zeng et al., 2020), a large-scale collection of Chinese medical dialogues crawled from the online consultation platform haodf.com \cite{he2020meddialog}. The corpus spans from 2010 to 2020, providing ID, URL, hospital, department, and description (including description, dialogue, Diagnosis, and suggestions, etc.) for each case. All patient names were de-identified. The dialogues span 29 broad clinical specialties(e.g., internal medicine, pediatrics, dentistry) and 172 fine-grained specialties (e.g., including cardiology, neurology, gastroenterology, urology). For our experiments, we use only the 2020 subset of MedDialog-CN, comprising 69,550 cases.

\subsubsection{2) Baselines and LLMs}

We compare the evaluated LLMs against three non-LLM baselines. 
First, the majority-class baseline assigns the most frequent reference 
grade in the evaluation set to all instances, providing a simple lower-bound reference. Second, the standard-based lookup baseline assigns sensitivity grades according to predefined mappings between information types and grades, without considering surrounding dialogue context. This baseline approximates a mention-level grading strategy in which the sensitivity of an entity is determined primarily from its category or subtype. Third, the deterministic context-rule baseline extends the lookup approach with manually specified contextual rules for explicit cues such as negation, uncertainty, test-result polarity, and selected granularity distinctions. This baseline provides a transparent comparison for assessing whether LLMs offer additional benefits beyond fixed context-sensitive rules.

We evaluate multiple contemporary LLMs representing different model families, model scales, and deployment settings. The final model set is selected before the formal evaluation and includes strong general-purpose proprietary models, models with strong language capabilities, and smaller or deployable models. The exact model versions, providers, access methods, and evaluation dates are recorded and kept fixed throughout the study to ensure reproducibility.

\subsubsection{3) Implementation Details}

All experiments follow the same frozen task definition and operational grading framework described in Section~3.1. For each target instance, the model is provided with the target span and its reference information type/subtype. Two input conditions are evaluated. In the mention-only condition, the model receives only the target span and information type, whereas in the full-context condition, the surrounding dialogue context is additionally provided. Except for the presence or absence of contextual information, the prompt structure and grading instructions are kept identical across the two conditions.

The prompt instructs each model to return a structured output containing four fields: assertion status, experiencer, applicability, and sensitivity grade. Assertion status is selected from \{confirmed, suspected, negated, hypothetical, planned, generic, unclear\}, while the final grade is selected from \{L1, L2, L3, L4, L5, N/A\}. The prompt explicitly specifies that predefined special-disease mentions can receive Level~5 only when the corresponding condition is confirmed, whereas suspected, negated, hypothetical, or planned mentions should not be assigned Level~5. No chain-of-thought reasoning is requested; an optional brief rationale is retained only for subsequent error analysis.

All models are evaluated using the same grading instructions. The temperature is set to zero, or to the lowest value supported by the corresponding provider, to reduce sampling variability. Other inference parameters, including top-$p$, maximum output length, and random seed where available, are also recorded for reproducibility.

For contrastive evaluation, the two members of each minimal pair are submitted independently and in randomized order so that the model is not informed of the paired relationship. The formal evaluation set is not used for prompt development or rule refinement. If few-shot examples are included, they are drawn exclusively from the pilot set and are fixed before formal evaluation. Model responses, model-version information, timestamps, parsing status, and retry records are retained for subsequent analysis. A single retry is permitted only when the initial response violates the required output format; responses that remain invalid after the retry are recorded as invalid outputs.

\subsubsection{4) Evaluation Metrics}

The evaluation metrics are organized according to the four research questions.

First, contextual gain is quantified by comparing each LLM under the
mention-only and full-context conditions. We report
$\Delta$Macro-F1 and $\Delta$QWK, defined as the differences in Macro-F1 and quadratic weighted kappa (QWK) between the full-context and mention-only settings for the same model. Positive values indicate improved grading performance when contextual information is provided.

Second, naturalistic grading performance is evaluated using Macro-F1, QWK, critical under-grading, and false-L5 assignment rate. Macro-F1 measures balanced classification performance across sensitivity grades, while QWK captures the ordinal agreement between predicted and protocol-based reference grades. Critical under-grading is defined as the proportion of reference-L5 instances predicted as Level~1--3. The false-L5 assignment rate is defined as the proportion of suspected, negated, hypothetical, or planned special-disease instances that are incorrectly predicted as Level~5.

Third, responsiveness to controlled contextual changes is assessed using pairwise exact accuracy and contrastive relation accuracy. Pairwise exact accuracy requires both members of a minimal pair to receive their correct reference labels. Contrastive relation accuracy evaluates whether the predicted relationship between the two members---higher, lower, same, or N/A---matches the reference relationship defined by the frozen grading protocol.

Fourth, model consistency is evaluated using repeated-run exact agreement and paraphrase consistency. Repeated-run exact agreement measures whether a model produces the same structured prediction across repeated evaluations of identical inputs, whereas paraphrase consistency assesses whether semantically equivalent reformulations receive consistent predictions.

\subsection{Preliminary Observations}
Our previous end-to-end extraction experiments showed relatively high aggregate grading performance, but qualitative error analysis revealed systematic failures at contextual boundaries. In particular, uncertain or ruled-out special-disease mentions were sometimes assigned Level 5, sensitive positive results were missed, and information granularity such as exact versus coarse dates was inconsistently interpreted. These observations motivated the present study to decouple entity extraction from sensitivity grading and directly evaluate contextual reasoning.

\section{Discussion}

\subsection{Context Matters for Sensitive-Information Grading}

The central premise of this study is that the sensitivity of information in online medical dialogues cannot always be determined from entity mentions alone. In particular, the presence of a highly sensitive disease term does not necessarily indicate that the corresponding condition has been established for a specific individual. The same disease entity may appear in confirmed, suspected, negated, hypothetical, or planned contexts, each of which carries a different semantic implication for privacy-sensitive grading.

This distinction is particularly important for high-sensitivity information. A mention-level approach that maps predefined disease terms directly to Level~5 may systematically over-grade suspected or explicitly ruled-out conditions. Conversely, failure to recognize a confirmed disclosure or a confirmatory positive test result may lead to serious under-grading. These two types of errors have different practical implications: false Level~5 assignments may unnecessarily restrict data access and reduce data utility, whereas severe under-grading may result in insufficient protection for genuinely sensitive
information.

The comparison between mention-only and full-context settings is therefore designed to quantify the extent to which surrounding dialogue contributes to reliable sensitivity grading. If contextual information improves Macro-F1 and ordinal agreement while reducing false Level~5 assignments and critical under-grading, this would suggest that sensitivity grading should be treated as a contextual reasoning task rather than a static entity-to-level mapping. Conversely, limited contextual gains would indicate that a substantial portion of the task can be addressed through simpler information-type mappings or deterministic rules.

The contrastive evaluation further isolates specific sources of contextual difficulty. Negation and uncertainty test whether models correctly distinguish confirmed disclosures from non-confirmed mentions, while experiencer contrasts assess whether models identify whose information is being described. Granularity contrasts examine whether models respond appropriately to differences in information specificity, such as full names versus surnames or exact dates versus coarse temporal references. These controlled comparisons provide a complementary perspective to aggregate performance on naturally occurring dialogues.

\subsection{Implications for Health-Data Governance}

The proposed framework has implications for the practical use of automated classification and grading systems in health-data governance. Existing regulatory standards provide an important basis for identifying health-data categories and general sensitivity levels, but free-text medical dialogues introduce semantic conditions that cannot always be represented by static lookup tables. Context-aware grading can therefore complement standard-informed mappings by incorporating assertion status, experiencer, result status, and information granularity into the grading process.

Importantly, this study does not treat LLM predictions as legally authoritative privacy decisions. The reference labels are operationalized from a transparent and predefined grading protocol, rather than assumed to represent a universal ground truth. In practice, an automated grading system could serve as a decision-support component for data governance, identifying potentially sensitive information and flagging ambiguous or high-risk cases for further review.

The comparison with transparent non-LLM baselines is also relevant to system deployment. Standard-based lookup provides a simple and interpretable solution for information types with relatively fixed mappings, whereas deterministic context rules may be sufficient for explicit linguistic cues such as simple negation. LLMs would be most valuable if they provide additional gains in more complex contexts, including long-range dependencies, colloquial expressions, or combinations of multiple contextual cues. Thus, the practical value of LLMs should be evaluated not only by overall accuracy but also by whether they offer measurable improvements over simpler and more interpretable alternatives.

For privacy-sensitive applications, error direction is as important as overall performance. A model with high aggregate accuracy may still be unsuitable for automated use if it frequently under-grades confirmed high-sensitivity information or systematically assigns Level~5 to non-confirmed disease mentions. Reporting critical under-grading and false Level~5 assignment rates alongside conventional classification metrics therefore provides a more operationally meaningful assessment of model reliability.

\section{Conclusion}
This study reframes sensitive-information grading in online medical consultations as a context-aware task rather than a direct mapping from entity mentions to fixed sensitivity levels. We introduce a standard-informed operational framework that explicitly incorporates assertion status, experiencer, result status, and information granularity, together with naturalistic and contrastive evaluation settings for current LLMs. By separating entity identification from contextual grading, the proposed design enables a more direct assessment of whether LLMs can distinguish sensitive mentions from contextually established disclosures. Future work will complete the planned multi-model evaluation and further examine the robustness and generalizability of the framework across datasets and clinical domains.

\printbibliography

\end{document}